\documentclass{article}
\usepackage{spconf,amsmath,graphicx}
\usepackage{url}
\usepackage{xcolor}
\usepackage{booktabs}
\usepackage{multirow}
\usepackage{amssymb}

\title{COMPRESSAI-VISION: OPEN-SOURCE SOFTWARE TO EVALUATE \\ COMPRESSION METHODS FOR COMPUTER VISION TASKS}
\name{Hyomin Choi,$^{\ast}$ Heeji Han,$^{\dag}$ Chris Rosewarne$^{\ddag}$ and Fabien Racap\'e$^{\ast}$}
\address{$^{\ast}$InterDigital, USA \hspace{0.5cm} $^{\dag}$Hanbat National University, South Korea \hspace{0.5cm} $^{\ddag}$Canon, Australia \\
}

\begin{document}
%
\maketitle
\begin{abstract}
With the increasing use of neural network (NN)-based computer vision applications that process image and video data as input, interest has emerged in video compression technology optimized for computer vision tasks. In fact, given the variety of vision tasks, associated NN models and datasets, a consolidated platform is needed as a common ground to implement and evaluate compression methods optimized for downstream vision tasks. \emph{CompressAI-Vision} is introduced as a comprehensive evaluation platform where new coding tools compete to efficiently compress the input of vision network while retaining task accuracy in the context of two different inference scenarios: ``remote'' and ``split'' inferencing. Our study showcases various use cases of the evaluation platform incorporated with standard codecs (under development) by examining the compression gain on several datasets in terms of bit-rate versus task accuracy. This evaluation platform has been developed as open-source software and is adopted by the Moving Pictures Experts Group (MPEG) for the development the Feature Coding for Machines (FCM) standard. The software is available publicly at {\small \url{https://github.com/InterDigitalInc/CompressAI-Vision}}.

\end{abstract}

\section{Introduction}
\label{sec:intro}
Driven by advances in deep learning methods and parallel computing capabilities, NN-based computer vision models are finding their way into a multitude of applications and are becoming mainstream consumers of images and video data. Due to the high computational complexity of the NN models, some vision applications process compressed image and video data, captured and transmitted from user devices, on the cloud server with high computational power. For other use cases, computationally optimized NN models on the neural processing unit (NPU)-enabled end device should be capable of processing a whole pipeline locally. Furthermore, recent research explores scenarios where a complex NN model is split into two parts to perform inference collaboratively and remotely. In this case, the intermediate output of the first split part is compressed and transmitted to the second part to accomplish the downstream vision task. In real-world applications, the first two scenarios have been widely adopted while the split-inference paradigm arose recently~\cite{kang2017neurosurgeon}. In either case, the NN models take compressed and transmitted data as input to accomplish downstream tasks for practical considerations. Nevertheless, there are few open-source evaluation platforms that allow the examination of such different inference scenarios end-to-end while applying compression to input of various computer vision models.

\begin{figure}[tbp]
  \centering
  \includegraphics[width=0.5\textwidth]{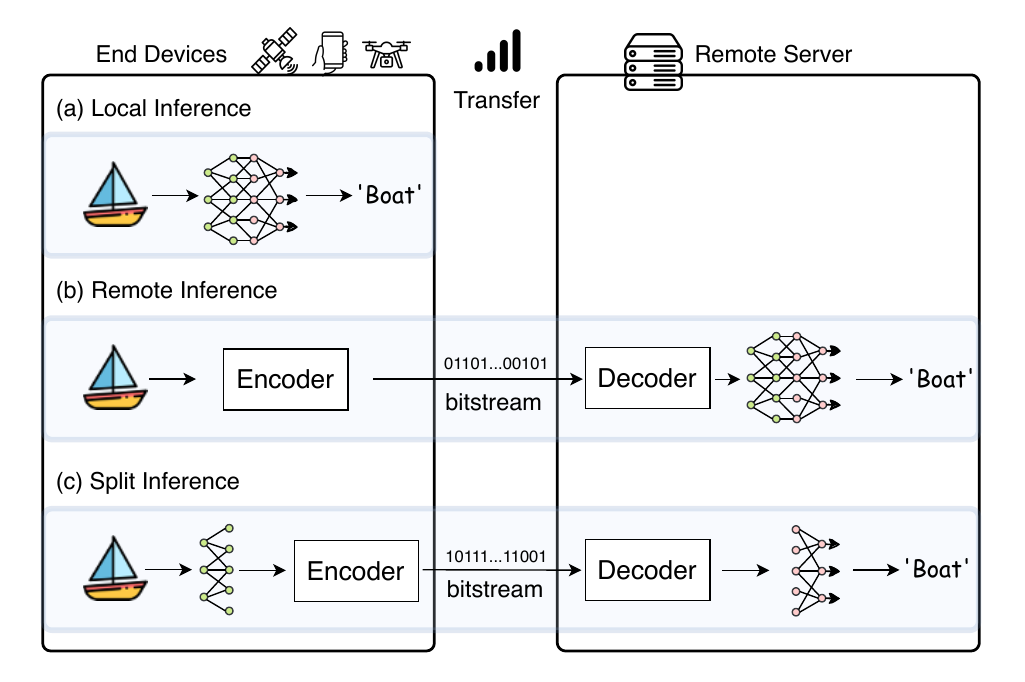}
  \vspace{-0.5cm}
  \caption{Three inference scenarios supported by CompressAI-Vision to evaluate compression performance for computer vision tasks: (a) requires no compression and is considered as a benchmark with default task accuracy, (b) compresses visual input data and a whole NN model runs on the server, and (c) compresses the intermediate feature data from the first split part, and the remaining is computed on the server.}
  \vspace{-0.5cm}
  \label{fig:three_inference_scenarios}
\end{figure}

This study introduces a comprehensive evaluation platform, called \emph{CompressAI-Vision}. This platform manages all aspects of the process, such as handling input formats, providing various computer vision models, providing a choice of codecs, and computing evaluation metrics, across the assessment pipeline. CompressAI-Vision is in use throughout Feature coding for machines standard development in the MPEG~\cite{fcm_cttc}, comparing the performance of different coding tools within aforementioned inference scenarios in terms of bitrate versus task accuracy. 

\begin{figure*}[tbp]
  \centering
  \includegraphics[width=0.85\textwidth]{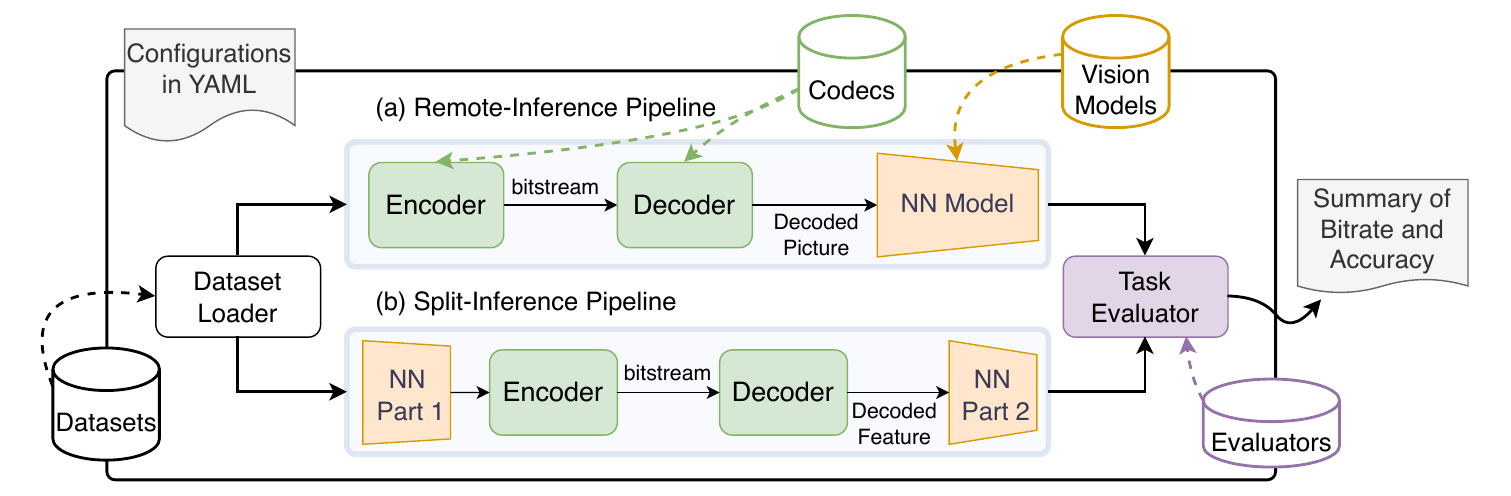}
  \vspace{-0.2cm}
  \caption{A brief schematic of CompressAI-Vision Platform that supports (a) remote-inference pipeline and (b) split-inference pipeline. The choice of pipeline along with specific codec, NN-based vision model, evaluator, and dataset is configurable via YAML file.}
  \vspace{-0.4cm}
  \label{fig:compressai_vision_platform}
\end{figure*}

\section{PRELIMINARY}
\label{sec:preliminaries}
The most straightforward inference scenario is the on-device processing, referred to as ``local inference,'' in which the captured scene is used directly as input\footnote{However, for practical reasons, most off-the-shelf vision models are trained on compressed images such as JPEG.} to an NN model powered by parallel computing, as shown in Fig.~\ref{fig:three_inference_scenarios}(a). One challenge is that available NN models by size, energy consumption, task accuracy, etc., vary depending on resource accessibility, which is a factor driving the cost of the devices in a wide range. In fact, moving beyond inference environments constrained by edge device resources, there exist two alternative inference scenarios, broadly categorized with a flexible form~\cite{su2022ai}: remote and split inference, which require compression and transmission.

\subsection{Remote Inference}
\label{ssec:}
Fig.~\ref{fig:three_inference_scenarios}(b) presents the remote inference scenario where the boat image is encoded and transmitted to the remote server with powerful computing resources. The decoded image is then fed into the NN model as input to identify the object `Boat.' From the machine's point of view, object-representing features are likely more important than the overall quality of reconstructed pixels. Early studies~\cite{choi2018high, galteri2018video, fischer2021saliency} demonstrated that vision task accuracy improves by compressing the machine vision focus area with a smaller quantization step size using conventional standard codecs, despite the reduced peak signal-to-ratio (PSNR). This scenario allows end-user devices to be less dependent on parallel computing power.

\subsection{Split Inference}
\label{ssec:split_inference}
For the case that pixel reconstruction is unnecessary, it should be more efficient in terms of bitrate to compress and transmit task-relevant information rather than to compress an entire picture. Furthermore, the server workload would benefit from offloading part of the NN model computation to the end-user device, especially when a huge number of inference requests are made. For example, as shown in Fig.~\ref{fig:three_inference_scenarios}(c), a network can be split into two parts: the first part runs on the source input, and the second part runs on the remote server. This strategy can alleviate the computational burden on the server while reducing transmission bandwidth. To incorporate the two parts remotely, the intermediate output of the first part should be compressed from a rate-accuracy perspective since the data volume of the uncompressed intermediate data is often much greater than the input. In both remote and split inference scenarios, inference output `Boat' could be feedback to the user if needed.

All three inference scenarios could be interchangeable depending on given environments and resource constraints. Essentially, compression plays a key role in establishing efficient remote and split inference scenarios. However, achieving efficient compression of these data for machine vision consumption remains a challenge.

\section{CompressAI-Vision THE EVALUATION PLATFORM}
\label{sec:compressai-vision}

\begin{table*}[t]
\centering
\caption{Summary of supported vision models with the list of split-point options. The last column provides tags for available split points, where the number of intermediate feature tensors at each split point is indicated in parentheses. Different tags split the associated model at various branches in the network architecture.}
\vspace{0.2cm}
\label{tbl:vision_models}
\resizebox{\linewidth}{!}{%
\setlength{\tabcolsep}{4pt}
\renewcommand{\arraystretch}{0.5} 
\tiny
\begin{tabular}{@{}cccc@{}}
\toprule
\textbf{Library}             & \textbf{Vision Model}                 & \textbf{Task}                                           & \textbf{Splits}                     \\ \midrule
\multirow{4}{*}{Detectron2~\cite{wu2019detectron2}} & faster\_rcnn\_R\_50\_FPN\_3x~\cite{ren2016faster}          & \multirow{2}{*}{Object Detection}                        & \multirow{2}{*}{"r2"(1), "c2"(1), "fpn"(4)}                   \\
                             & faster\_rcnn\_X\_101\_32x8d\_FPN\_3x~\cite{ren2016faster}  &                                                          &                   \\ \cmidrule(l){2-4} 
                             & mask\_rcnn\_R\_50\_FPN\_3x~\cite{he2017mask}            & \multirow{2}{*}{Object Detection, Instance Segmentation} & \multirow{2}{*}{"r2"(1), "c2"(1), "fpn"(4)}                   \\
                             & mask\_rcnn\_X\_101\_32x8d\_FPN\_3x~\cite{he2017mask}    &                                                          &                   \\ \midrule
JDE~\cite{wang2019towards}                          & jde\_1088x608~\cite{wang2019towards}                         & Multiple People Tracking                                 & "dn53"(3), "alt1"(3) \\ \midrule
YOLOX~\cite{yolox2021}                        & yolox\_darknet53~\cite{yolox2021}                      & Object Detection                                         & "l13"(1), "l37"(1)                        \\ \midrule
MMPOSE~\cite{mmpose2020}                       & rtmo\_multi\_person\_pose\_estimation~\cite{lu2023rtmo} & Pose Estimation                                          & "backbone"(2), "neck"(2)                          \\ \bottomrule
\end{tabular}}
\end{table*}
CompressAI-Vision is a python-implemented evaluation platform in which the three inference pipelines can be examined with various configurable parts, as shown in Fig.~\ref{fig:compressai_vision_platform}. First, there are dataset loaders that handle two formats (e.g., image and video) and cooperate with various task annotations in different styles such as COCO~\cite{COCO}, VOC~\cite{imagenet2015}. However, the library does not include the datasets; these must be prepared separately by users. For a sequence of frames (i.e., a video) as input to vision models that process frame by frame, it is necessary to separate the video into individual frame files. When necessary, the pre-processing can include the color space conversion of the input frame from YUV420 to RGB, as typically used in computer vision networks. The appropriate dataset loader should be chosen based on the vision model used for evaluation within the selected inference pipeline.

Table~\ref{tbl:vision_models} shows several vision models supported in the latest version of CompressAI-Vision, along with their split-point options. For the fundamental vision like object detection and instance segmentation, 
we support most well-known NN-models from Detectron~\cite{wu2019detectron2} and YOLOX~\cite{yolox2021}. Also, the models ~\cite{wang2019towards, lu2023rtmo} for multiple people tracking and pose estimation are integrated. As an open-source software library, the integration of new vision models into CompressAI-Vision is always available through public contribution and a proper review process. 

The use of the remote-inference pipeline ignores the split point. However, for the split-inference pipeline, it must be set to compress and transmit one or multiple coded feature tensors at the split point for the downstream vision task. By analyzing the network architecture of each vision model, several split points are pre-selected, and its interfaces are designed to be used with the tags. For example, a common split point in R-CNN networks is at the output of the feature pyramid network (FPN), from which four feature tensors at different scales are coded. Further details about each split point can be found in the Github repository\footnote{\url{https://github.com/InterDigitalInc/CompressAI-Vision}}.

\begin{figure}[t]
    \centering
    \begin{minipage}[b]{0.32\linewidth}
    \centering
    \includegraphics[width=\textwidth]{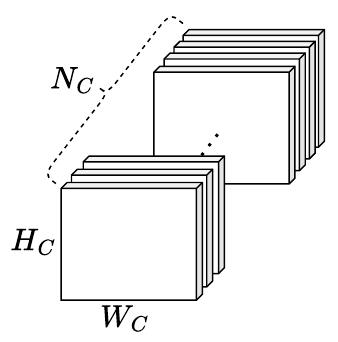}
    \centerline{(a)}\medskip
    \end{minipage}
    \hspace{1cm}
    \centering
    \begin{minipage}[b]{0.45\linewidth}
    \centering
    \includegraphics[width=\textwidth]{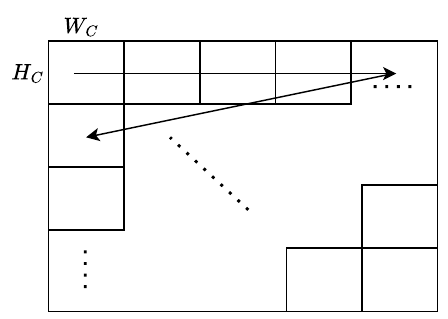}
    \centerline{(b)}\medskip
    \end{minipage}
\vspace{-0.5cm}
\caption{An example of (a) a 3D tensor of feature channels and (b) a 2D frame of the tiled feature channels.}
\vspace{-0.5cm}
\label{fig:tensor_and_tile}
\end{figure}

Different input characteristics for the codec, such as integer pixels vs. computed (floating-point) features or a 2D frame vs. a 3D feature tensor with $N_C \times H_C \times W_C$, where $N_C$ is the number of feature channels and $H_C \times W_C$ is the channel resolution as depicted in Fig.~\ref{fig:tensor_and_tile}(a), necessitate different compression optimization strategies depending on the inference pipeline. 
In essence, the reference software of standard codecs like AVC~\cite{avc_std}, HEVC~\cite{hevc_std}, VVC~\cite{vvc_std} straightforwardly cooperates with the remote-inference pipeline. You may also use other codecs with FFmpeg ~\cite{tomar2006converting}, which is an option available on the platform. Within the split-inference pipeline, the 3D feature tensor must be properly reshaped to interface with existing video codecs, which take 2D frames as input. By default, the platform reshapes the 3D tensor into a 2D frame using tiled feature channels arranged in raster scan order, as shown in Fig.~\ref{fig:tensor_and_tile}(b); however, other methods can also be exercised.

The use of existing codecs may not lead to optimal results in terms of bitrate vs. task accuracy, since these codecs have been designed and optimized for human vision perception over decades. In order to meet the recent need for new codecs optimized for machine vision consumption, MPEG has been developing new standard compression methods for machines: Video Coding for Machines (VCM) and Feature Coding for Machines (FCM). The CompressAI-Vision platform is capable of evaluating reference software for both of these standards under development following their respective common test conditions (CTC). In particular, this platform has served MPEG FCM as the common evaluation platform by supporting all the use cases described in their CTC~\cite{fcm_cttc}.

Finally, there are multiple task evaluators that come with the vision models and are integrated into the evaluation platform. For vision models supporting tasks like object detection, instance segmentation, multiple people tracking, and pose estimation, the relevant evaluator must be selected in the yaml configuration to properly examine the task accuracy using associated metrics such as mean average precision (mAP)~\cite{COCO,imagenet2015} and multi-object tracking accuracy (MOTA)~\cite{bernardin2008evaluating}.

\section{USE FOR STANDARD DEVELOPMENT IN MPEG}
\label{sec:use_for_standard_development_in_mpeg}

CompressAI-Vision serves as the common evaluation platform for the under-developing MPEG VCM and MPEG FCM. This platform is capable of benchmarking the codecs under various common test conditions~\cite{fcm_cttc, vcm_ctc}. In addition, it integrates a visualization tool to enable an intuitive analysis of the trade-off between compression gain and task accuracy~\cite{vis_tool}.

This section showscases CompressAI-Vision, which examines and compares the coding performance of the codecs across various settings. Since each codec pursues different optimalities across two different inference pipelines, an immediate comparison is challenging. Nevertheless, thanks to the common platform, a consistent experimentation is available across the four inference scenarios: local inference, remote inference, remote inference with MPEG VCM, and split inference with MPEG FCM.

\vspace{-0.3cm}
\subsection{Experiment Setup} 
\label{ssec:evaluation_setup_across_standards}
To compare the coding efficiency of MPEG VCM and MPEG FCM, their respective reference software, the VCM Reference Software (VCM-RS) and the Feature Compression Test Model (FCTM), interface with CompressAI-Vision. The codecs are then cross-compared, each being evaluated under the test conditions of the other including coding configurations. Moreover, the standard codecs~\cite{avc_std, hevc_std, vvc_std} are examined by replacing the VCM-RS and the FCTM in the associated inference pipeline as benchmarks. We also benchmark the task accuracy with compression compared to the original task performance with the local-inference pipeline, identifying the compression impact to end task results.

We borrowed three conditions of random access (RA), low delay (LD) and all-intra (AI) with end-to-end (e2e) configuration from VCM CTC~\cite{vcm_ctc} and the LD configuration from FCM CTTC~\cite{fcm_cttc} for the cross-comparison. For the FCM CTTC, the intra-refresh frame is inserted periodically depending on the frame rate of the source video for the LD configuration, corresponding to the intra period applied in the RA configuration of the VCM CTC.

\subsection{Datasets and Metrics} 
\label{ssec:datasets}

\begin{table}[t]
\centering
\caption{Evaluation datasets for MPEG FCM and VCM}
\vspace{0.2cm}
\label{tbl:dataset_summary}
\resizebox{\linewidth}{!}{%
\renewcommand{\arraystretch}{0.5} 
\scriptsize
\begin{tabular}{@{}cccccc@{}}
\toprule
\textbf{Dataset}                    & \begin{tabular}[c]{@{}c@{}}\textbf{Data} \\ \textbf{Type}\end{tabular}   & \begin{tabular}[c]{@{}c@{}}\textbf{\# of Images} \\ \textbf{Sequences}\end{tabular} & \textbf{Task}                                                             & \textbf{FCM} & \textbf{VCM} \\ \midrule
\multirow{2}{*}{\begin{tabular}[c]{@{}c@{}}Open \\ Images~\cite{openimages}\end{tabular}} & \multirow{3}{*}{Image} & 5000                                                                     & \begin{tabular}[c]{@{}c@{}}Instance \\ Segmentation\end{tabular} & \checkmark &     \\ \cmidrule(l){3-6} 
                           &                        & 5000                                                                     & Object Detection                                                 & \checkmark &     \\ \cmidrule(r){1-1} \cmidrule(l){3-6} 
FLIR~\cite{flir}                       &                        & 300                                                                      & Object Detection                     &  & \checkmark \\ \midrule                           
SFU-HW-Obj~\cite{sfu-hw-obj-v32,choi2021dataset}                & \multirow{3}{*}{Video} & 14                                                                       & Object Detection                                                 & \checkmark & \checkmark \\ \cmidrule(r){1-1} \cmidrule(l){3-6} 
TVD~\cite{tvd}                        &                        & 3                                                                        & Object Tracking                                                  & \checkmark & \checkmark \\ \cmidrule(r){1-1} \cmidrule(l){3-6} 
HiEVE~\cite{lin2020human}                      &                        & 5                                                                        & Object Tracking                                                  & \checkmark &     \\ \bottomrule
\end{tabular}}
\vspace{-0.5cm}
\end{table}
A number of datasets listed in Table~\ref{tbl:dataset_summary} are adopted to evaluate the FCTM and VCM-RS in the context of MPEG standard development. From the OpenImages dataset~\cite{openimages}, 5,000 images were specifically chosen for Instance Segmentation, and another 5,000, with a few thousand being common to the Instance Segmentation set, were selected for Object Detection. Also, a small dataset, FLIR~\cite{flir}, consisting of 300 infrared images is adopted for the object detection task. There are three video datasets, each of which includes several sequences of various lengths. For example, the TVD and the HiEVE include three and five video sequences, respectively, for Object Tracking, evaluated with the MOTA metric~\cite{bernardin2008evaluating}. SFU-HW-Obj~\cite{choi2021dataset} provides the object labels with bounding boxes on 14 raw video sequences. The MPEG FCM and VCM adopt different subset of the datasets as marked in Table~\ref{tbl:dataset_summary}. In the experiments, we used the common datasets between them for the cross-evaluation.



\subsection{Cross-Evaluation of FCTM and VCM-RS}
\label{ssec:performance_comparison}
First, FCTM v6.1\footnote{\url{https://git.mpeg.expert/MPEG/Video/fcm/fctm}} and VCM-RS v0.12\footnote{\url{https://git.mpeg.expert/MPEG/Video/VCM/VCM-RS}} are evaluated under the test condition of VCM CTC~\cite{vcm_ctc} as summarized in Table~\ref{tbl:FCTM_under_VCMCTC}, with quantization parameter (QP) adjustment to ensure comparable performance in task accuracy. 
For the object detection task on the SFU-HW-Obj dataset, the FCTM saves bits 79.35\% and 69.02\% for Class C and Class D, respectively, compared to VCM-RS at the same task accuracy. On average, the FCTM decrease bits by -58.33\%, -41.43\%, and -72.70\% compared to the VCM-RS results under the RA, LD, and AI with e2e configurations, respectively, in terms of BD-Bitrate. In contrast, for the object tracking task on the TVD dataset, the FCTM increases the bits by 26.63\% and 18.69\% compared to the VCM-RS under RA and LD with e2e configuration, particularly due to the degraded performance on the TVD-01 sequence.

\begin{table}[t]
\centering
\caption{BD-Bitrate comparison of FCTM v6.1 against VCM-RS v0.12 under the VCM CTC~\cite{vcm_ctc}}
\vspace{0.2cm}
\label{tbl:FCTM_under_VCMCTC}
\resizebox{\linewidth}{!}{%
\renewcommand{\arraystretch}{0.7} 
\scriptsize
\begin{tabular}{@{}ccrrr@{}}
\toprule
\multicolumn{2}{c}{\textbf{Dataset}} & \textbf{RA (e2e)} & \textbf{LD (e2e)} & \textbf{AI (e2e)} \\ \midrule
\multirow{5}{*}{SFU-HW-Obj~\cite{choi2021dataset}} & Class A & 7.83\% & 21.67\% & -26.25\% \\
                        & Class B & -38.12\% & -11.54\% & -54.69\% \\
                        & Class C & -79.35\% & -74.58\% & -88.12\% \\
                        & Class D & -69.02\% & -46.48\% & -82.41\% \\
                        & Average & -58.33\% & -41.43\% & -72.70\% \\ \midrule
\multicolumn{2}{c}{\textbf{Dataset}} & \textbf{RA (e2e)} & \textbf{LD (e2e)} & \textbf{AI (e2e)} \\ \midrule
\multirow{8}{*}{TVD~\cite{tvd}}    & TVD-01-1 & 45.06\% & 56.69\% & -1.87\% \\
                        & TVD-01-2 & 63.48\% & 35.05\% & N/A \\
                        & TVD-01-3 & 134.84\% & 13.08\% & 17.80\% \\
                        & TVD-02-1 & -8.53\% & 68.35\% & 1.10\% \\
                        & TVD-03-1 & -23.45\% & -40.05\% & 1.68\% \\
                        & TVD-03-2 & -14.71\% & -20.15\% & 3.31\% \\
                        & TVD-03-3 & -10.27\% & 17.84\% & 8.42\% \\
                        & Average  & 26.63\% & 18.69\% & N/A \\ \bottomrule
\end{tabular}}
\end{table}

\begin{figure}[t!]
    \centering
    \begin{minipage}[b]{1\linewidth}
    \centering
    \centerline{\includegraphics[width=7.5cm]{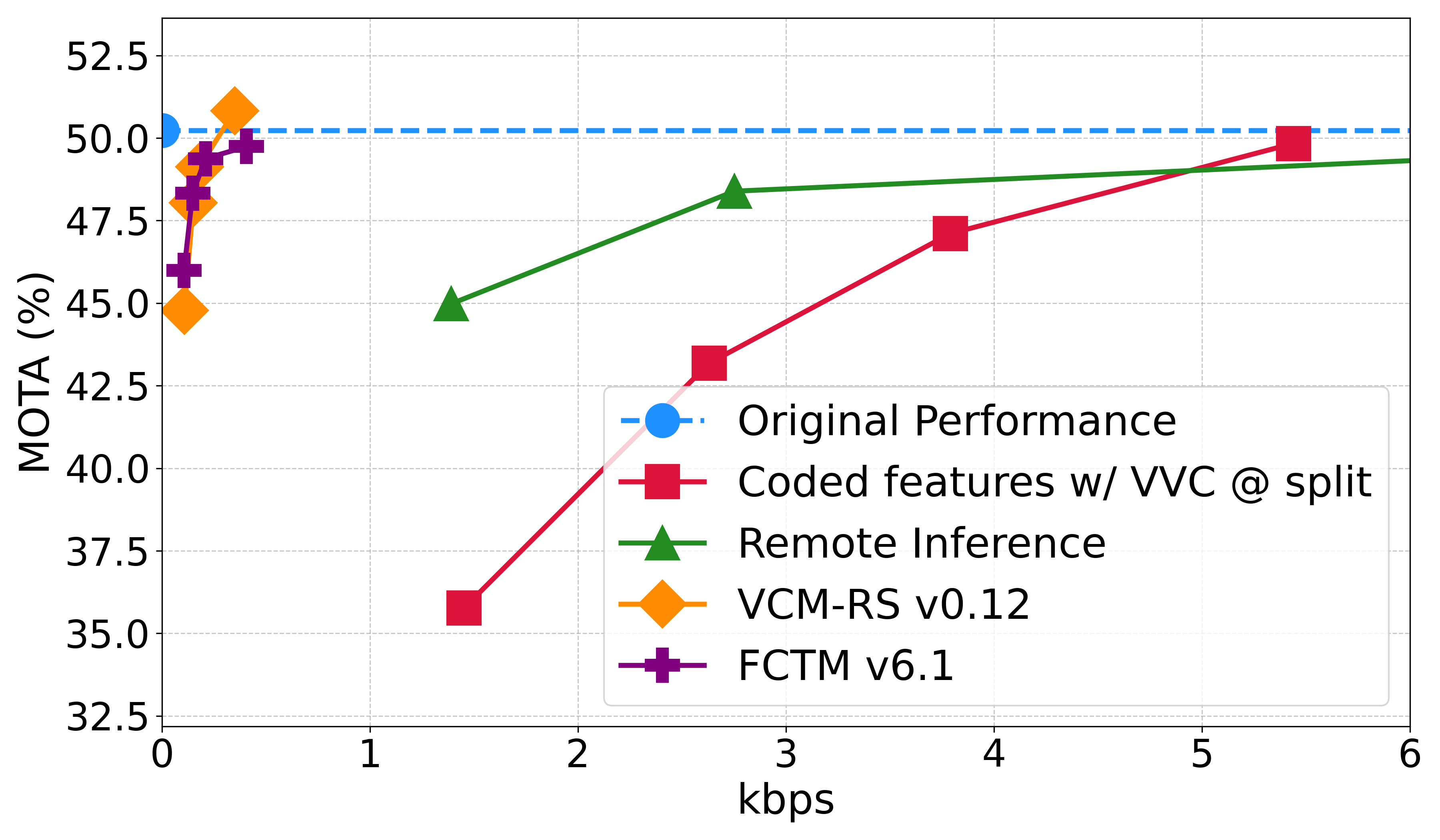}}
    \centerline{(a) TVD}\medskip
    \end{minipage}
    \begin{minipage}[b]{1\linewidth}
    \centering
    \centerline{\includegraphics[width=7.5cm]{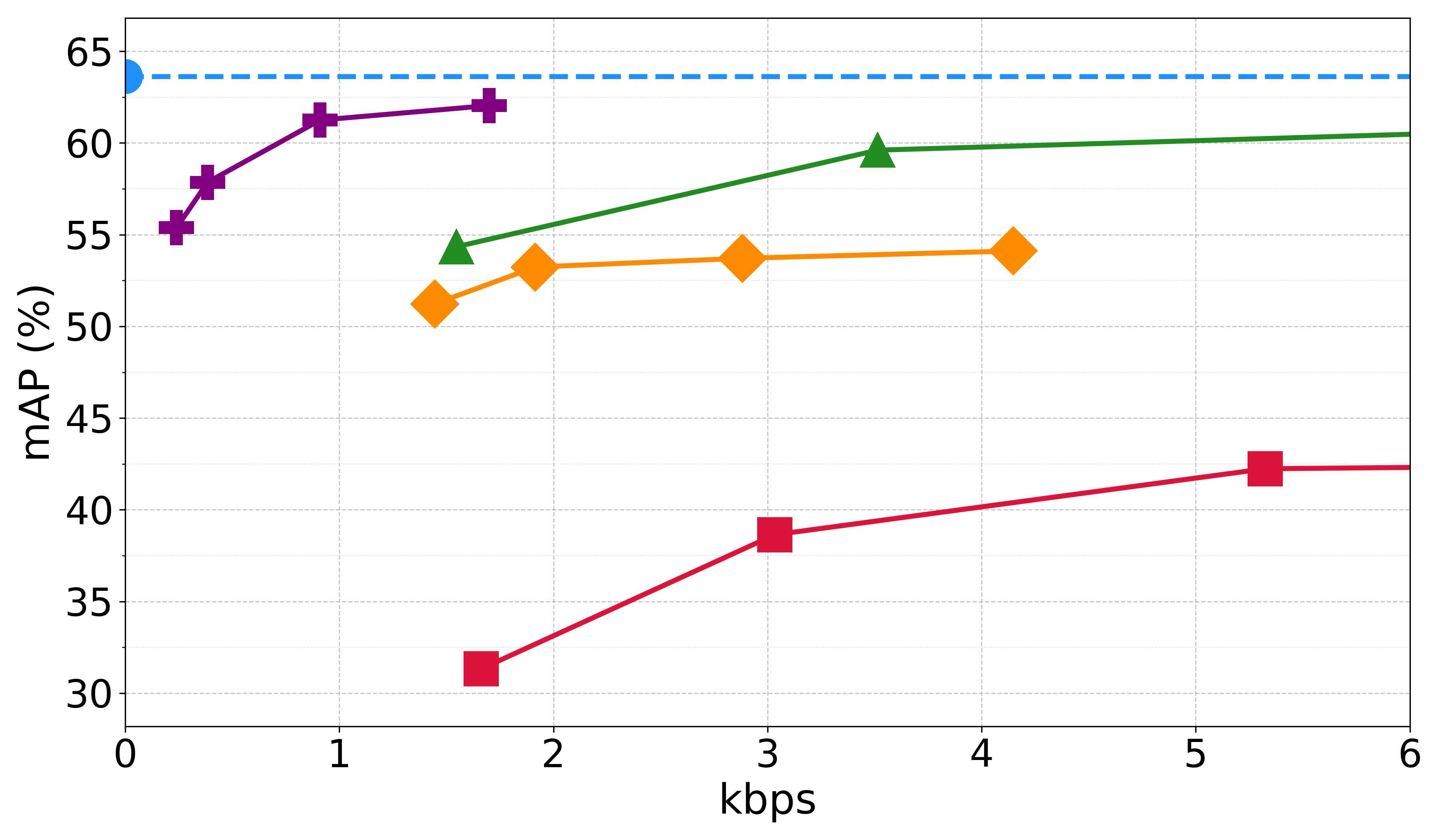}}
    \centerline{(b) SFU-HW-Obj ClassC}\medskip
    \end{minipage}
    \vspace{-0.5cm}
    \caption{Rate-Accuracy curves of several compression methods with different inference scenarios under the FCM CTTC~\cite{fcm_cttc} evaluated on (a) TVD and (b) SFU-HW-Obj ClassC. The same legend applies to both figures.}
    \vspace{-0.3cm}
\label{fig:VCM_under_FCMCTTC}
\end{figure}
Conversely, VCM-RS v0.12 was evaluated under the FCM CTTC~\cite{fcm_cttc} using the LD configuration, and the summarized Rate-Accuracy (RA) curves are shown in Fig.~\ref{fig:VCM_under_FCMCTTC}. On the TVD dataset in Fig.~\ref{fig:VCM_under_FCMCTTC}(a), the VCM-RS and FCTM substantially outperform other methods, including remote inference (green) and the compression of multiple features with VVC at the JDE~\cite{wang2019towards} split (red). Also, both the VCM-RS and the FCTM reaches near-lossless accuracy at their highest bitrate\footnote{Compression noise often improves the original accuracy randomly as shown in the yellow curve.} The performance of compression methods for machines is often contingent on the specific datasets and the computer vision networks to evaluate. For instance, Fig.~\ref{fig:VCM_under_FCMCTTC}(b) illustrates the RA curves on the SFU-HW-Obj ClassC dataset. FCTM achieves near-lossless accuracy and demonstrates a superior RA gain when compared to methods for remote inference scenarios and direct feature compression using VVC.

Overall, these experiments showcases that CompressAI-Vision is designed to support cross-evaluation by enabling consistent testing of different coding tools across heterogeneous inference pipelines and model configurations.

\subsection{Comparison of FCTM with Different Inner Codecs}
Because FCM is designed to work with other inner codecs such as AVC~\cite{avc_std} and HEVC~\cite{hevc_std}, CompressAI-Vision's feature of supporting a range of conventional video codecs throughout its inference pipelines naturally complements FCM. Table~\ref{tbl:BD_rate_HM_JM} compares the coding performance of using JM-19.1\footnote{\url{https://vcgit.hhi.fraunhofer.de/jvet/JM/-/tree/JM-19.1}} and HM-18.0\footnote{\url{https://vcgit.hhi.fraunhofer.de/jvet/HM/-/tree/HM-18.0}} to using VTM-23.3\footnote{\url{https://vcgit.hhi.fraunhofer.de/jvet/VVCSoftware_VTM/-/tree/VTM-23.3}} as the inner codec for FCTM v6.1, with the same QP applied in all cases.
The evaluation was conducted using a subset of sequences from the SFU-HW-Objects dataset.
Although VTM demonstrates superior performance compared to alternative inner codecs in FCTM, the Bitrate-vs-Accuracy performance gap between VTM and HM or JM is considerably smaller than the typical performance gap observed in terms of Bitrate-vs-PSNR. Ultimately, CompressAI-Vision effectively manages different inner codec configurations, facilitating consistent and comparative coding performance evaluations.

\begin{table}[tb!]
\centering
\caption{Performance comparison of FCTM v6.1 using HM-18.0 and JM-19.1 as inner codec against FCTM v6.1 with VTM-23.3}
\vspace{0.2cm}
\label{tbl:BD_rate_HM_JM}
\tiny
\resizebox{0.9\linewidth}{!}{%
\renewcommand{\arraystretch}{0.1} 
\begin{tabular}{@{}llll@{}}
\toprule
\multicolumn{2}{c}{Dataset} & HM-18.0 & JM-19.1 \\ \midrule
\multirow{3}{*}{SFU-HW-Obj} & Class AB & 4.90\%     & 39.40\%     \\ \cmidrule(l){2-4} 
                            & Class C  & 12.86\%     & 53.28\%     \\ \cmidrule(l){2-4} 
                            & Class D  & 12.42\%     & 38.55\%     \\ \bottomrule
\vspace{-0.2cm}
\end{tabular}}
\end{table}

\section{FUTURE WORKS}
\label{sec:future_works}
To leverage cutting-edge vision models, the platform will support vision transformer~\cite{dosovitskiy2020image}-based architectures. This enables to explore the impact of compression noise on the embedding space in relation to the downstream task. Moreover, exploring a multi-task network is encouraged to understand the interplay between compression and different tasks, ultimately leading to optimal coding methods. This could also lead to new compression paradigms optimized for parallel processing of various machine vision tasks.

\section{CONCLUSIONS}
\label{sec:conclusions}

CompressAI-Vision enables a comprehensive study of compression methods for various computer vision tasks within local, remote, and split inference frameworks, all on a common evaluation platform. Beyond automated calculations in numerical metrics, it also supports visual comparison. The platform's active use in MPEG for codecs under development further demonstrates its benefit and practicality as an evaluator. As open-source software, CompressAI-Vision offers inherent scalability and extendability, fostering community contributions and future advancements.

%
%
%

\vfill\pagebreak

\bibliographystyle{IEEEbib_abbrev}
\small
\bibliography{refs}

\end{document}